%% file: paper.tex
\pdfoutput=1
\documentclass[10pt,twocolumn,letterpaper]{article}

\usepackage{cvpr}
\usepackage{times}
\usepackage{epsfig}
\usepackage{graphicx}
\usepackage{amsmath}
\usepackage{amssymb}
\usepackage{multirow}
\usepackage{diagbox}
\usepackage{amsmath,booktabs}
\usepackage{bm}
\usepackage[flushleft]{threeparttable}
\usepackage[flushleft]{threeparttable}
\usepackage{comment}
\usepackage{chngpage}
\usepackage{array}

\usepackage[pagebackref=true,breaklinks=true,letterpaper=true,colorlinks,bookmarks=false]{hyperref}

\cvprfinalcopy % *** Uncomment this line for the final submission

\def\cvprPaperID{1248} % *** Enter the CVPR Paper ID here

\ifcvprfinal\fi

\begin{document}

%%%%%%%%% TITLE
\title{PU-Net: Point Cloud Upsampling Network}

\author{First Author\\
Institution1\\
Institution1 address\\
{\tt\small firstauthor@i1.org}
% For a paper whose authors are all at the same institution,
% omit the following lines up until the closing ``}''.
% Additional authors and addresses can be added with ``\and'',
% just like the second author.
% To save space, use either the email address or home page, not both
\and
Second Author\\
Institution2\\
First line of institution2 address\\
{\tt\small secondauthor@i2.org}
}

\author{Lequan Yu\thanks{Equal contribution.}${\ \ }^{1,3}$\quad Xianzhi Li$^{* 1}$\quad Chi-Wing Fu$^{1,3}$\quad Daniel Cohen-Or$^2$\quad Pheng-Ann Heng$^{1,3}$\\
	%$^1$Department of Computer Science and Enginering \hspace{10mm} $^2$School of Computer Science,\\
	\hspace{-10mm}$^1$The Chinese University of Hong Kong \hspace{10mm} $^2$ Tel Aviv University\\
	 $^3$Guangdong Provincial Key Laboratory of Computer Vision and Virtual Reality Technology,\\
	 Shenzhen Institutes of Advanced Technology, Chinese Academy of Sciences, China\\
	\hspace{-10mm}{\tt\small \{lqyu,xzli,cwfu,pheng\}@cse.cuhk.edu.hk}\hspace{10mm}{\tt\small dcor@mail.tau.ac.il}\qquad
}

\newcommand{\TODO}[1]{{\color{red}{[TODO: #1]}}}
\newcommand{\phil}[1]{{\color[rgb]{0.3,0.7,0.3}{[PH: #1]}}}
\newcommand{\leqn}[1]{{\color{blue}{[LQ: #1]}}}
\newcommand{\xz}[1]{{\color[rgb]{1.0,0.6,0}{[XZ: #1]}}}
\newcommand{\para}[1]{\vspace{.05in}\noindent\textbf{#1}}
\def\ie{\emph{i.e.}}
\def\eg{\emph{e.g.}}
\def\etal{{\em et al.}}
\def\etc{{\em etc.}}

\maketitle

\input{abstract}
\input{introduction}

\input{method}

\input{training}
\input{experiments}
\input{conclusion}

\para{Acknowledgments.}\
We thank anonymous reviewers for the comments and suggestions.
The work is supported in part by the National Basic Program of China, the 973 Program (Project No. 2015CB351706), the Research Grants Council of the Hong Kong Special Administrative Region (Project no. CUHK 14225616), the Shenzhen Science and Technology Program (No. JCYJ20170413162617606), and the CUHK strategic recruitment fund.

{\small
	\bibliographystyle{ieee}
	\bibliography{paper}
}

\input{supp}

\end{document}

%% file: abstract.tex
\begin{abstract}
	Learning and analyzing 3D point clouds with deep networks is challenging due to the sparseness and irregularity of the data.
	In this paper, we present a data-driven point cloud upsampling technique.
	The key idea is to learn multi-level features per point and expand the point set via a multi-branch convolution unit implicitly in feature space. The expanded feature is then split to a multitude of features, which are then reconstructed to an upsampled point set.
	Our network is applied at a patch-level, with a joint loss function that encourages the upsampled points to remain on the underlying surface with a uniform distribution.
	We conduct various experiments using synthesis and scan data to evaluate our method and demonstrate its superiority over some baseline methods and an optimization-based method.  Results show that our upsampled points have better uniformity and are located closer to the underlying surfaces.
\end{abstract}

%% file: introduction.tex
\section{Introduction}

Point cloud is a fundamental 3D representation that has drawn increasing attention due to the popularity of various depth scanning devices.
Recently, pioneering works~\cite{qi2016pointnet,qi2017pointnet++,klokov2017escape} began to explore the possibility of reasoning point clouds by means of deep networks for understanding geometry and recognizing 3D structures.
In these works, the deep networks directly extract features from the raw 3D point coordinates without using traditional features, \eg, normal and curvature. These works present impressive results for 3D object classification and semantic scene segmentation.

In this work we are interested in an upsampling problem: 
given a set of points, generate a denser set of points to describe the underlying geometry by learning the geometry of a training dataset. 
This upsampling problem is similar in spirit to the image super-resolution problem~\cite{shi2016real,ledig2016photo}; however, dealing with 3D points rather than a 2D grid of pixels poses new challenges.
First, unlike the image space, which is represented by a regular grid, point clouds do not have any spatial order and regular structure.
Second, the generated points should describe the underlying geometry of a latent target object, meaning that they should roughly lie on the target object surface.
%, relying a sparse set of input points only.
%%
Third, the generated points should be informative and should not clutter together.
% near the input points.
Having said that, the generated output point set should be more uniform on the target object surface.
Thus, simple interpolation between input points cannot produce satisfactory results.

To meet the above challenges, we present a data-driven point cloud upsampling network.
Our network is applied at a patch-level, with a joint loss function that encourages the upsampled points to remain on the underlying surface with a uniform distribution.
The key idea is to learn multi-level features per point, and then {\em expand the point set via a multi-branch convolution unit implicitly in feature space\/}. The expanded feature is then split to a multitude of features, which are then reconstructed to an upsampled point set.

Our network, namely PU-Net, learns geometry semantics of point-based patches from 3D models, and applies the learned knowledge to upsample a given point cloud.
It should be noted that unlike previous network-based methods designed for 3D point sets~\cite{qi2016pointnet,qi2017pointnet++,klokov2017escape}, the number of input and output points in our network are {\em not the same\/}.

We formulate two metrics, distribution uniformity and distance deviation from underlying surfaces, to quantitatively evaluate the upsampled point set, and test our method using variety of synthetic and real-scanned data. We also evaluate the performance of our method, and compare it to baseline and state-of-the-art optimization-based methods. Results show that our upsampled points have better uniformity, and are located closer to the underlying surfaces.

\if 0
In the following we first revised related works. 
Then in Section~\ref{sec:architecture}, our PU-Net architecture will be introduced in detail with four different components.
We will further elaborate the patch-based training manner and introduce the novel joint loss function for end-to-end training in Section~\ref{sec:training}. 
The formulation of evaluation metrics and a variety of experiments will be presented in Section~\ref{sec:experiments}.
Finally, Section~\ref{sec:conclusion} concludes the paper.
\fi

\begin{figure*}[!t]
	\centering
	\includegraphics[width=1.0\linewidth]{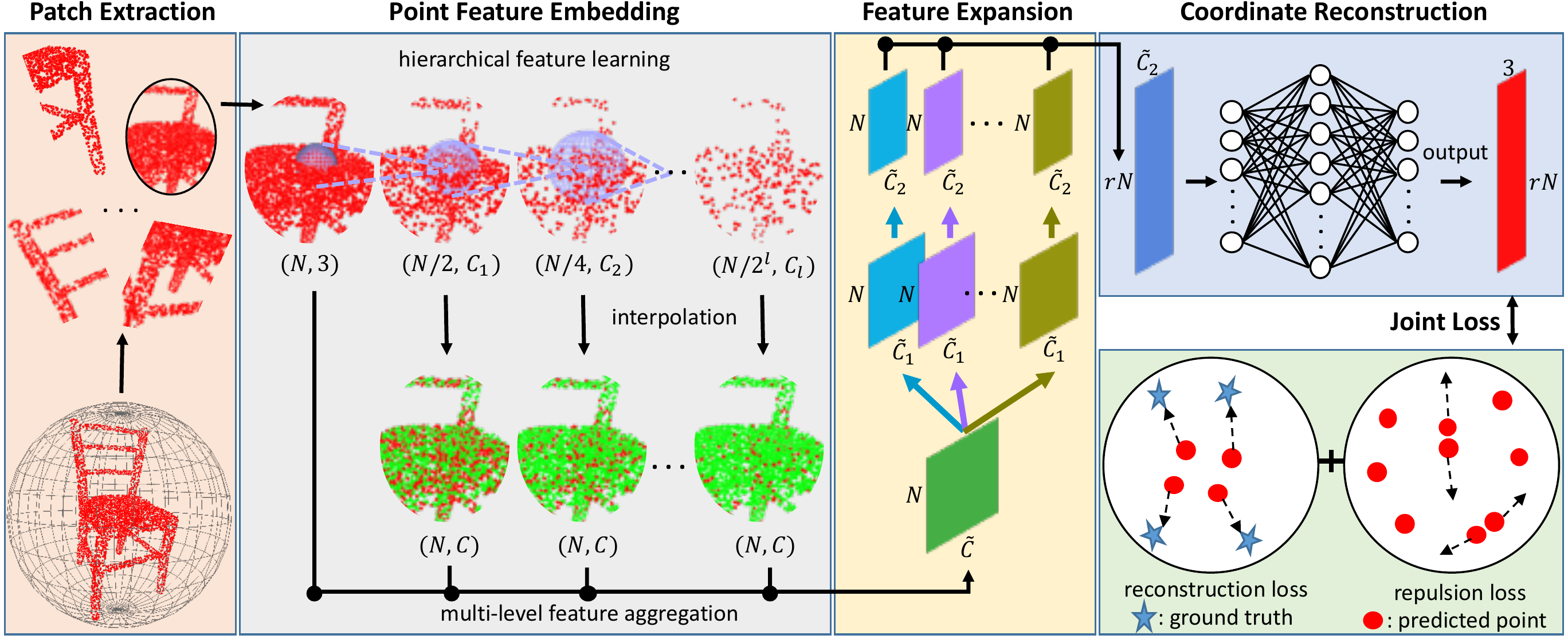}\vspace{1mm}
	\caption{The architecture of PU-Net (better view in color). The input has $N$ points, while the output has $rN$ points, where $r$ is the upsampling rate. $C_i$, $\tilde{C}$ and $\tilde{C}_i$ represent the feature channel number. We restore different level features for the original $N$ points with interpolation, and reduce all level features to a fixed dimension $C$ with a convolution. The red color in the point feature embedding component shows the original and progressively subsampled points in hierarchical feature learning, while the green color indicates the restored features. 
		We jointly adopt the reconstruction loss and repulsion loss in the end-to-end training of PU-Net.}
	\label{fig:arch}\vspace{-2mm}
\end{figure*}

\vspace{-7pt}
\paragraph{Related work: optimization-based methods.}
An early work by Alexa~\etal~\cite{alexa2003computing} upsamples a point set by interpolating points at vertices of a Voronoi diagram in the local tangent space.
%%, where the main purpose is to enhance the point set for rendering.
%%
Lipman~\etal~\cite{lipman2007parameterization} present a locally optimal projection (LOP) operator for points resampling and surface reconstruction based on an $L_1$ median.
The operator works well even when the input point set contains noise and outliers.
Successively, Huang~\etal~\cite{huang2009consolidation} propose an improved weighted LOP to address the point set density problem.

Although these works have demonstrated good results, they make a strong assumption that the underlying surface is smooth, thus restricting the method's scope.
Then, Huang~\etal~\cite{huang2013edge} introduce an edge-aware point set resampling method by first resampling away from edges and then progressively approaching edges and corners. 
However, the quality of their results heavily relies on the accuracy of the normals at given points and careful parameter tuning. 
It is worth mentioning that Wu~\etal~\cite{wu2015deep} propose a deep points representation method to fuse consolidation and completion in one coherent step.
Since its main focus is on filling large holes, global smoothness is, however, not enforced, so the method is sensitive to large noise.
Overall, the above methods are not data-driven, thus heavily relying on priors.

\vspace{-7pt}
\paragraph{Related work: deep-learning-based methods.}
%%{\em Deep-learning-based methods.} \
%%
Points in a point cloud do not have any specific order nor follow any regular grid structure, so only a few recent works adopt a deep learning model to directly process point clouds.
Most existing works convert a point cloud into some other 3D representations such as the volumetric grids~\cite{maturana2015voxnet,wu20153d,riegler2016octnet,dai2017shape} and geometric graphs~\cite{bruna2013spectral,masci2015geodesic} for processing.
Qi~\etal~\cite{qi2016pointnet,qi2017pointnet++} firstly introduced a deep learning network for point cloud classification and segmentation; in particular, the PointNet++ uses a hierarchical feature learning architecture to capture both local and global geometry context.
Subsequently, many other networks were proposed for high-level analysis problems with point clouds~\cite{klokov2017escape,hua2017pointwise,li2018pointcnn,wang2017sgpn,qi2017frustum}.
However, they all focus on global or mid-level attributes of point clouds.
In another work, Guerrero~\etal~\cite{guerrero2017pcpnet} developed a network to estimate the local shape properties in point clouds, including normal and curvature.
Other relevant networks focus on 3D reconstruction from 2D images~\cite{fan2016point,lin2017learning,Groueix2018AtlasNet}.
To the best of our knowledge, there are no prior works focusing on point cloud upsampling.

\if 0
In some other aspects, Klokov~\etal~\cite{klokov2017escape} propose the Kd-network for unstructured point cloud recognition. 
%%
The recent work of Fan~\etal~\cite{fan2016point} is most related to us. They introduce a prediction network to regress a point cloud from a given single image.
Since the input modality is an image, it can employ a convolutional neural network to extract features.
%%
Consequently, Lin~\etal~\cite{lin2017learning} present a pseudo-renderer, as an approximation of true rendering operation, to generate a 3D point cloud from a single image. Achlioptas~\etal~\cite{achlioptas2017representation} introduce a deep auto-encoder network for processing 3D point clouds.
To the best of our knowledge, there are no prior works focusing on the point cloud upsampling problem.
\fi

%% file: method.tex
\section{Network Architecture}
\label{sec:architecture}

%We take a patch-based approach to train the network and learn geometry semantics.
Given a 3D point cloud with point coordinates in nonuniform distributions, our network aims to output a denser point cloud that follows the underlying surface of the target object while being uniform in distribution.
Our network architecture (see Fig.~\ref{fig:arch}) has four components: \emph{patch extraction}, \emph{point feature embedding}, \emph{feature expansion}, and \emph{coordinate reconstruction}.
First, we extract patches of points in varying scales and distributions from a given set of prior 3D models (Sec.~\ref{sec:patch}).
Then, the \emph{point feature embedding} component maps the raw 3D coordinates to a feature space by hierarchical feature learning and multi-level feature aggregation (Sec.~\ref{sec:featureEmbed}).
After that, we expand the number of features using the \emph{feature expansion} component (Sec.~\ref{sec:featureShuffle}) and reconstruct the 3D coordinates of the output point cloud via a series of fully connected layers in the \emph{coordinate reconstruction} component (Sec.~\ref{sec:coordConstruct}).
%%

%##################################################

\subsection{Patch Extraction}
\label{sec:patch}
We collect a set of 3D objects as prior information for training.
These objects cover a rich variety of shapes, from smooth surface to shapes with sharp edges and corners.
Essentially, for our network to upsample a point cloud, it should learn local geometry patterns from the objects.
This motivates us to take a patch-based approach to train the network and to learn the geometry semantics.

In detail, we randomly select $M$ points on the surface of these objects.
From each selected point, we grow a surface patch on the object, such that any point on the patch is within a certain geodesic distance ($d$) from the selected point over the surface.
Then, we use Poisson disk sampling to randomly generate $\hat{N}$ points on each patch as the referenced ground truth point distribution on the patch.
In our upsampling task, both local and global context contribute to a smooth and uniform output.
Hence, we set $d$ with varying sizes, so that we can extract patches of points on the prior objects with varying scale and density.

%##################################################

\subsection{Point Feature Embedding}
\label{sec:featureEmbed}
To learn both local and global geometry context from the patches, we consider the following two feature learning strategies, whose benefits complement each other:

\para{Hierarchical feature learning.}
Progressively capturing features of growing scales in a hierarchy has been proved to be an effective strategy for extracting local and global features.
Hence, we adopt the recently proposed hierarchical feature learning mechanism in PointNet++~\cite{qi2017pointnet++} as the very frontal part in our network. 
To adopt hierarchical feature learning for point cloud upsampling, we specifically use a relatively small grouping radius in each level, since generating new points usually involves more of the local context than the high-level recognition tasks in~\cite{qi2017pointnet++}.

\para{Multi-level feature aggregation.}
Lower layers in a network generally correspond to local features in smaller scales, and vice versa.
For better upsampling results, we should optimally aggregate features in different levels. 
Some previous works adopt skip-connections for cascaded multi-level feature aggregation~\cite{long2015fully,ronneberger2015u,qi2017pointnet++}.
However, we found by experiments that such top-down propagation is not very efficient for aggregating features in our upsampling problem.
Therefore, we propose to directly combine features from different levels and let the network learn the importance of each level~\cite{hariharan2015hypercolumns,xie2015holistically,hou2016deeply}.

Since the input point set on each patch (see point feature embedding in Fig.~\ref{fig:arch}) is subsampled gradually in hierarchical feature extraction, we concatenate the point features from each level by first restoring the features of all original points from the downsampled point features by the interpolation method in PointNet++~\cite{qi2017pointnet++}.
Specifically, the features of an interpolated point $x$ in level $\ell$ is calculated by:
\begin{equation}
f^{(\ell)}(x)  = \frac{\sum_{i=1}^3w_i(x)f^{(\ell)}(x_i)}{\sum_{i=1}^3w_i(x)} \ ,
%f^{(\ell)}(x)  = 1 \ ,
\end{equation}
%\phil{something strange in the above equation... $f^{(\ell)}(x)$ appears in the middle?}
%%
where $w_i(x)$$=$$1/d(x,x_i)$, which is an inverse distance weight, and $x_i$, $x_2$, $x_3$ are the three nearest neighbors of $x$ in level $\ell$.
We then use a 1$\times$1 convolution to reduce the interpolated feature in different level to the same dimension, \ie, $C$. 
Finally, we concatenate the features from each level as the embedded point feature $f$. 

\subsection{Feature Expansion}
\label{sec:featureShuffle}
After the point feature embedding component, we expand the number of features in the feature space.
This is equivalent to expanding the number of points, since {\em points and features are interchangeable\/}.
Suppose the dimension of $f$ is $N \times \tilde{C}$, $N$ is the number of input points and $\tilde{C}$ is the feature dimension of the concatenated embedded feature.
The \emph{feature expansion} operation would output a feature $f'$ with dimension $rN \times \tilde{C}_2$, where $r$ is the upsampling rate and $\tilde{C}_2$ is the new feature dimension.  
Essentially, this is similar to feature upsampling in image-related tasks, which can be done by deconvolution~\cite{long2015fully} (also known as transposed convolution) or interpolation~\cite{dong2016image}. 
However, it is nontrivial to apply these operations to point clouds due to the non-regularity and unordered properties of points.

We therefore propose an efficient \emph{feature expansion} operation based on the sub-pixel convolution layer~\cite{shi2016real}. 
This operation can be represented as:
\begin{equation}
f' = \mathcal{RS}( \ [ \ \mathcal{C}_1^2(\mathcal{C}_1^1(f)) \ , \  ... \ , \ \mathcal{C}_r^2(\mathcal{C}_r^1(f)) \ ] \ ) \ ,
\end{equation}
where $\mathcal{C}_i^1(\cdot)$ and $\mathcal{C}_i^2(\cdot)$ are two sets of separate 1$\times$1 convolutions, and $\mathcal{RS}(\cdot)$ is a reshape operation to convert an $N\times r\tilde{C}_2$ tensor to a tensor of size $rN \times \tilde{C}_2$.
We emphasize that the feature in the embedding space has already encapsulated the relative spatial information from the neighborhood points via the efficient multi-level feature aggregation, so we do not need to explicitly consider the spatial information when performing this feature expansion operation. 

It is worth mentioning that the $r$ feature sets generated from the first convolution $\mathcal{C}_i^1(\cdot)$ in each set have a high correlation, and this would cause the final reconstructed 3D points to be located too close to one another. 
Hence, we further add another convolution (with separate weights) for each feature set. 
Since we train the network to learn the $r$ different convolutions for the $r$ feature sets, these new features can include more diverse information, thus reducing their correlations.
This \emph{feature expansion} operation can be implemented by applying separated convolutions to the $r$ feature sets; see Fig.~\ref{fig:arch}. 
It can also be implemented by more computation efficient grouped convolution~\cite{krizhevsky2012imagenet,xie2016aggregated,zhang2017shufflenet}. 

\subsection{Coordinate Reconstruction}
\label{sec:coordConstruct}
In this part, we reconstruct the 3D coordinates of output points from the expanded feature with the size of $rN \times \tilde{C}_2$. 
Specifically, we regress the 3D coordinates via a series of fully connected layers on the feature of each point, and the final output is the upsampled point coordinates $rN \times 3$.

%% file: training.tex
\section{End-to-End Network Training}
\label{sec:training}

%%%%%%%%%%%%%%%%%%%%%%%%%%%%%%%%%%%%%
\subsection{Training Data Generation}
\label{sec:strategy}
Point cloud upsampling is an ill-posed problem due to the uncertainty or ambiguity of upsampled point clouds.
Given a sparse input point cloud,  there are many feasible output point distributions. 
Therefore, we do not have the notion of ``correct pairs'' of input and ground truth.
To alleviate this problem, we propose an on-the-fly input generation scheme.
Specifically, the referenced ground truth point distribution of a training patch is fixed, 
whereas {\em the input points are randomly sampled from the ground truth point set with a downsampling rate of $r$\/} at each training epoch.
Intuitively, this scheme is equivalent to simulating many feasible output point distributions for a given sparse input point distribution. 
Additionally, this scheme can further enlarge the training dataset, allowing us to depend on a relatively small dataset for training.

%%%%%%%%%%%%%%%%%%%%%%%%%%%%%%%%%%%%
\subsection{Joint Loss Function}
\label{sec:objectFunc}
We propose a novel joint loss function to train the network in an end-to-end fashion.
%In order to train the network in an end-to-end fashion, we need to design an appropriate loss function to measure the quality of the output point cloud.
As we mentioned earlier, the function should encourage the generated points to be located on the underlying object surfaces in a more uniform distribution. 
Therefore, we design a joint loss function that combines the \emph{reconstruction loss} and \emph{repulsion loss}.

%%%%%%%%%%%%%%%%%%%%%%%%%%%%
\para{Reconstruction loss.}  To put points on the underlying object surfaces, we propose to use the Earth Mover's distance (EMD)~\cite{fan2016point} as our reconstruction loss to evaluate the similarity between the predicted point cloud $S_{p} \subseteq \mathbb{R}^3$ and the referenced ground truth point cloud $S_{gt}\subseteq \mathbb{R}^3$:
\begin{equation} 
\label{equ:rec}
	L_{rec} =d_{EMD}(S_{p}, S_{gt})=\min_{\phi:S_{p}\rightarrow S_{gt}} \sum_{x_i\in S_{p}} \|x_i-\phi(x_i)\|_2,
\end{equation}
where $\phi:S_{p} \rightarrow S_{gt}$ indicates the bijection mapping.

Actually, Chamfer Distance (CD) is another candidate for evaluating the similarity between two point sets.
However, compared with CD, EMD can better capture the shape (see~\cite{fan2016point} for more details) to encourage the output points to be located close to the underlying object surfaces. 
Hence, we choose to use EMD in our reconstruction loss.

%%%%%%%%%%%%%%%%%%%%%%%%%%%%%%
\para{Repulsion loss.} Although training with the reconstruction loss can generate points on the underlying object surfaces, the generated points tend to be located near the original points.
To distribute the generated points more uniformly, we design the repulsion loss, which is represented as:
\begin{equation}
\label{equ:rep}
L_{rep} = \sum_{i =0}^{\hat{N}}\sum_{i'\in K(i)}\eta(\|x_{i'}-x_i \|)w(\|x_{i'}-x_i\|) \ ,
\end{equation} 
where $\hat{N} = rN$ is the number of output points, $K(i)$ is the index set of the $k$-nearest neighbors of point $x_{i}$, and $\|\cdot\|$ is the L2-norm.
$\eta(r) = -r$ is called the \emph{repulsion term}, which is a decreasing function to penalize $x_{i}$ if $x_{i}$ is located too close to other points in $K(i)$.
To penalize $x_{i}$ only when it is too close to its neighboring points, we add two restrictions: 
(i) we only consider points $x_{i'}$ in the $k$-nearest neighborhood of $x_{i}$; and
(ii) we use the fast-decaying weight function $w(r)$ in the repulsion loss; that is, we follow~\cite{lipman2007parameterization,huang2009consolidation} to set $w(r) = e^{-r^2/h^2}$ in our experiments. 

Altogether, we train the network in an end-to-end manner by minimizing the following joint loss function:
\begin{equation}
\label{equ:objectFunc}
L(\bm{\theta}) = L_{rec} +\alpha L_{rep} +\beta\|\bm{\theta}\|^2,
\end{equation}
where $\bm{\theta}$ indicates the parameters in our network, $\alpha$ balances the reconstruction loss and repulsion loss, and $\beta$ denotes the multiplier of weight decay. 
For simplicity, we ignore the index of each training sample.

%% file: experiments.tex
\section{Experiments}
\label{sec:experiments}

%%%%%%%%%%%%%%%%%%%%%%%%%%%%%%%%%%%%%%%%%%%%%%%
\subsection{Datasets}
Since there are no public benchmarks for point cloud upsampling, we collect a dataset of 60 different models from the Visionair repository~\cite{timmurphy.org}, ranging from smooth non-rigid objects (\eg, Bunny) to steep rigid objects (\eg, Chair).
Among them, we randomly select 40 for training, and use the rest for testing\footnote{The complete object list can be found in the supplementary material.}.
We crop 100 patches for each training object, and we use $M$$=$$4000$ patches to train the network in total.
For testing objects, we use Monte-Carlo random sampling approach to sample 5000 points on each object as input.
To further demonstrate the generalization ability of our network, we directly test our well-trained network on the SHREC15~\cite{Lian2015} dataset, which contains 1200 shapes from 50 categories.
In detail, we randomly choose one model from each category for testing, considering that each category contains 24 similar objects in various poses. 
As for ModelNet40~\cite{wu20153d} and ShapeNet~\cite{chang2015shapenet}, we found it hard to extract patches from those objects due to the low mesh quality (\eg, holes, self-intersection, etc.).
Therefore, we use them for testing; see the supplementary material for the results.

%%%%%%%%%%%%%%%%%%%%%%%%%%%%%%%%%%%%%%%%%%%%%%%%
\subsection{Implementation Details}
\label{sec:implementationdetails}
The default point number $\hat{N}$ of each patch is 4096, and the upsampling rate $r$ is 4. Therefore, each input patch has 1024 points.
To avoid overfitting, we augment the data by randomly rotating, shifting and scaling the data.
We use 4 levels with grouping radii 0.05, 0.1, 0.2 and 0.3 in the point feature embedding component, and the dimension $C$ of the restored feature is 64.  For details on other network architecture parameters, please see our supplementary material. 
Parameters $k$ and $h$ in repulsion loss are set as 5 and 0.03, respectively.
The balancing weights $\alpha$ and $\beta$ are set as 0.01 and $\beta$ = $10^{-5}$, respectively.
The implementation is based on TensorFlow\footnote{ \url{https://github.com/yulequan/PU-Net}}.
For the optimization, we train the network for 120 epoch using the Adam~\cite{kingma2014adam} algorithm with a minibatch size of 28 and a learning rate of 0.001.
Generally, the training took about 4.5h on the NVIDIA TITAN Xp GPU. 

%##################################################################################################
\begin{figure}[!t]
	\centering
	\includegraphics[width=0.9\linewidth]{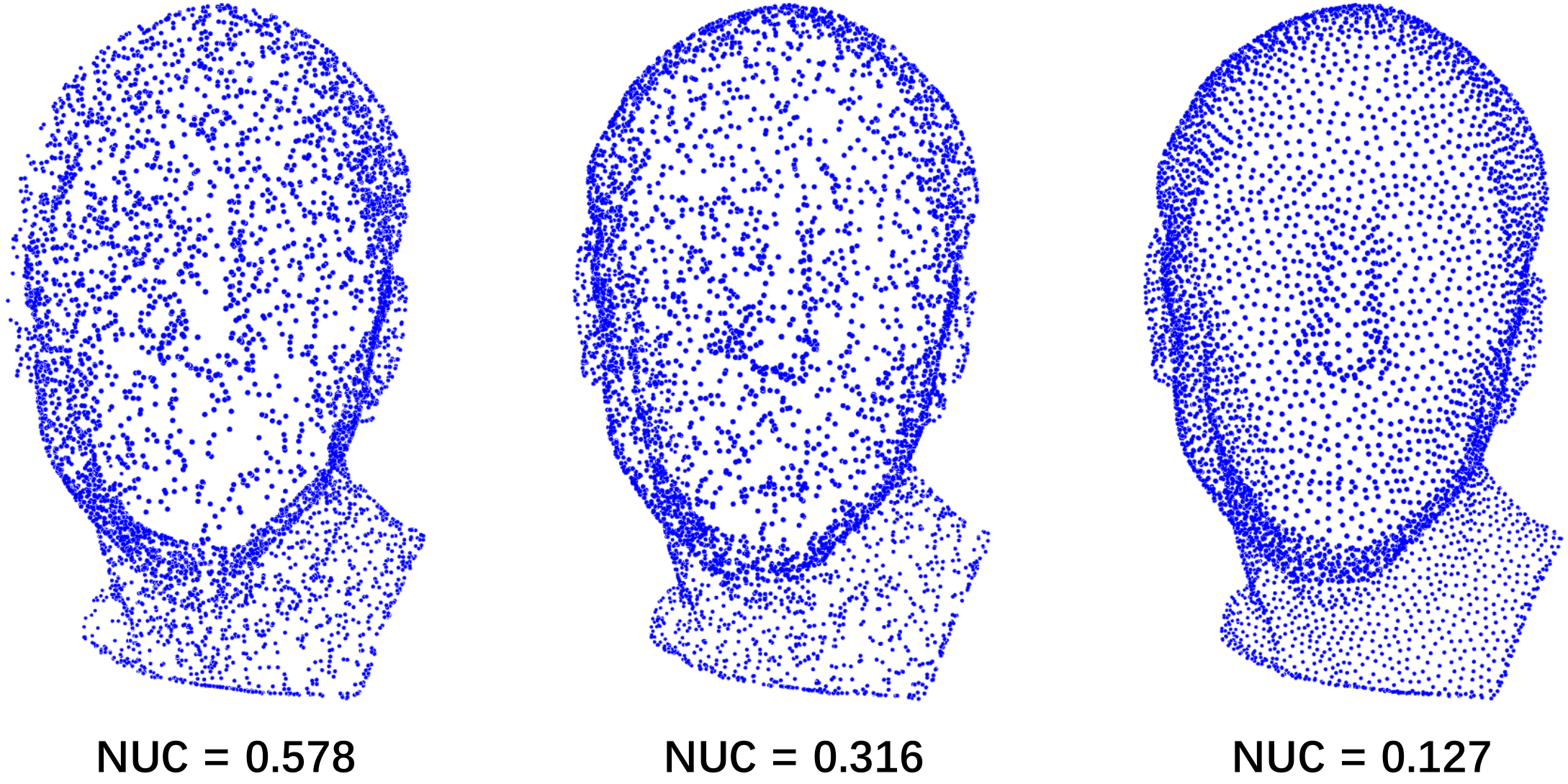}\vspace{1mm}
	\caption{Example point distributions with corresponding normalized uniformity coefficients (NUC) computed with $p=0.2\%$.}
	\label{fig:metric}\vspace{-2mm}
\end{figure}
%##################################################################################################

%%%%%%%%%%%%%%%%%%%%%%%%%%%%%%%%%%%%%%%%%%%%%%%%%
\subsection{Evaluation Metric}
\label{sec:metric}

To quantitatively evaluate the quality of the output point sets, we formulate two metrics to measure the deviation between the output points and the ground truth meshes, as well as the distribution uniformity of the output points.
For surface deviation, we find the closest point $\hat{x}_i$ on the mesh for each predicted point $x_i$, and calculate the distance between them. 
Then we compute the mean and standard deviation over all the points as one of our metrics. 

As for the uniformity metric, we randomly put $D$ equal-size disks on the object surface ($D = 9000$ in our experiments) and calculate the standard deviation of the number of points inside the disks. We further normalize the density of each object and then compute the overall uniformity of the point sets over all the objects in the testing dataset.
Therefore, we define the \emph{normalized uniformity coefficient} (NUC) with disk area percentage $p$ as:
\begin{equation}
\begin{split}
avg &= \frac{1}{K*D}\sum_{k=1}^{K}\sum_{i=1}^{D}\frac{n_i^k}{N^k*p},\\
NUC &= \sqrt{\frac{1}{K*D}\sum_{k=1}^{K}\sum_{i=1}^{D}(\frac{n_i^k}{N^k*p}-avg)^2},
\end{split}
\end{equation}
where $n_i^k$ is the number of points within the $i$-th disk of the $k$-th object, $N^k$ is the total number of points on the $k$-th object, $K$ is the total number of test objects, and $p$ is the percentage of the disk area over the total object surface area. 
Note that we use geodesic distance rather than Euclidean distance to form the disks.
Fig.~\ref{fig:metric} shows three different point distributions with their corresponding NUC values.
As we can see, the proposed NUC metric can effectively reveal the point set uniformity: the lower the UNC value, the more uniform the point set distribution is.
% is in accordance with human visual comparison.

%%%%%%%%%%%%%%%%%%%%%%%%%%%%%%%%%%%%%%%%%%%%%%%%%
\subsection{Comparisons with Other Methods}
\label{sec:comparison}

\newcommand{\BE}[1]{{\textbf{#1}}}
\begin{table*}
	\caption{Quantitative comparison on our collected dataset.}
	\label{tab:comparision1}
	\centering
	\begin{center}
		\begin{tabular}{l|c|c|c|c|c|c||c|c||c} \toprule[1pt]
			\multirow{2}*{Method}& \multicolumn{6}{c||}{NUC with different $p$} & \multicolumn{2}{c||}{Deviation ($10^{-2}$)}  & \multirow{2}*{Time (s)}\\
			\cline{2-9} 
			&0.2\%	&0.4\%	&0.6\%	&0.8\% 	&1.0\%	&1.2\%  	& mean  & std 	& \\ \hline \hline
			Input 		&0.316	&0.224	&0.185	&0.164	&0.150	&0.142		& -		&-		&-\\ \hline
			PointNet~\cite{qi2016pointnet}		&0.409	&0.334	&0.295	&0.270	&0.252	&0.239		&2.27	&3.18	&\BE{0.05}\\ \hline
			PointNet++~\cite{qi2017pointnet++}	&0.215	&0.178	&0.160	&0.150	&0.143	&0.139	&1.01	&0.83	&0.16\\ \hline
			PointNet++(MSG)~\cite{qi2017pointnet++}	&0.208	&0.169	&0.152	&0.143	&0.137	&0.134	&0.78	&0.61	&0.45\\ \hline
			PU-Net (Ours)			&\BE{0.174}&\BE{0.138}&\BE{0.122}&\BE{0.115}&\BE{0.112}&\BE{0.110}&\BE{0.63}	&\BE{0.53}	&0.15\\ \bottomrule[1pt]
		\end{tabular}
	\end{center}
	\vspace{-2.5mm}
\end{table*}

\begin{table*}
	\caption{Quantitative comparison on SHREC15 dataset.}
	\label{tab:comparision2}
	\centering
	\begin{center}
		\begin{tabular}{l|c|c|c|c|c|c||c|c||c} \toprule[1pt]
			\multirow{2}*{Method}& \multicolumn{6}{c||}{NUC with different $p$} & \multicolumn{2}{c||}{Deviation ($10^{-2}$)}  & \multirow{2}*{Time (s)}\\
			\cline{2-9} 
			&0.2\%	&0.4\%	&0.6\%	&0.8\% 	&1.0\%	&1.2\%  	& mean  & std 	& \\ \hline \hline
			Input 								&0.315	&0.222	&0.184	&0.165	&0.153	&0.146		& -		&-		&-\\ \hline
			PointNet~\cite{qi2016pointnet}		&0.490	&0.405	&0.360	&0.330	&0.309	&0.293		&2.03	&2.94	&\BE{0.05}\\ \hline
			PointNet++~\cite{qi2017pointnet++}	&0.259	&0.218	&0.197	&0.185	&0.176	&0.170		&0.88	&0.75	&0.16\\ \hline
			PointNet++(MSG)~\cite{qi2017pointnet++}	&0.250	&0.199	&0.175	&0.161	&0.153	&0.148	&0.69	&0.59	&0.45\\ \hline
			PU-Net (Ours)								&\BE{0.219}	&\BE{0.173}	&\BE{0.154}	&\BE{0.144}	&\BE{0.140}	&\BE{0.137} 		&\BE{0.52}	&\BE{0.45}	&0.15\\ \bottomrule[1pt]	
		\end{tabular}
	\end{center}
	\vspace{-4.5mm}
\end{table*}

\begin{figure}[t]
	\centering
	\includegraphics[width=1.0\linewidth]{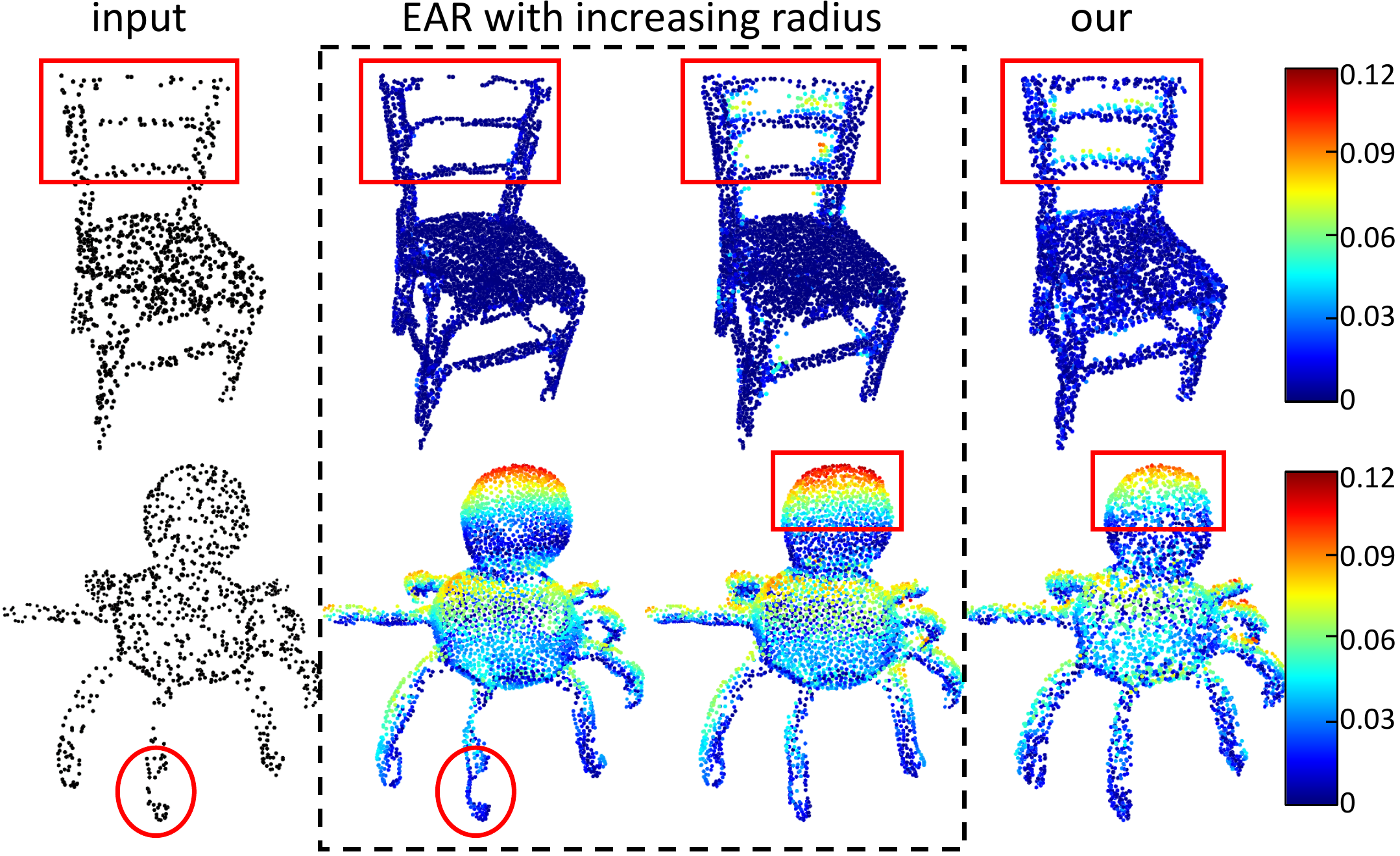}\vspace{1mm}
	\caption{Comparison with the EAR method~\cite{huang2013edge}.
We color-code all point clouds to show the deviation from the ground truth mesh.
For the EAR method, the radius of the Chair model is 0.1 and 0.182, while the radius of the Spider model is 0.106 and 0.155.}
	\label{fig:ear}
	\vspace{-2mm}
\end{figure}

\para{Comparison with an optimization-based method.} \
We compare our method with the Edge Aware Resampling (EAR) method~\cite{huang2013edge}, which is a state-of-art method for point cloud upsampling.
The results are shown in Fig.~\ref{fig:ear}, where the  Chair is from our collected testing dataset and the Spider is from SHREC15.
We color-code the point clouds to show the deviation from the ground truth meshes.
There are 1024 points in the input and we do a 4X upsampling.
Since EAR relies on the normal information, to be fair, we calculate the normal according to the ground truth mesh. 
We show two results of EAR with increasing radius, while setting other parameters to their default values.
As we can see, the radius parameter has a great influence on EAR's performance. 
For relatively small radius, the output has low surface deviation but the added points are not uniform, while more outliers are introduced if the radius is large. 
In contrast, our method can better balance the deviation and uniformity without the need to carefully tune the parameters. 
%\leqn{I am not sure whether the following argument is proper.}This comparison shows that the optimization-based EAR method has no ability to utilize the geometry semantics for upsampling, but this kind of information is very important for generating meaningful points.

%%%%%%%%%%%%%%%%%%%%%
\para{Comparison with deep learning-based methods.} 
As far as we know, we are not aware of any deep learning-based method for point cloud upsampling, so we design some baseline methods for comparison. 
Since PointNet~\cite{qi2016pointnet} and PointNet++~\cite{qi2017pointnet++} are pioneers for 3D point cloud reasoning with deep learning techniques, we design the baselines based on them.
%However, these two networks are designed for high level point cloud recognition problems rather than upsampling, therefore we adopt their semantic segmentation network architecture  for point feature embedding and use one set of convolutions to do feature expansion. 
Specifically, we adopt the semantic segmentation network architecture for point feature embedding and use one set of convolutions for feature expansion. 
Note that we consider two versions of PointNet++: basic PointNet++ and PointNet++ with multi-scale grouping (MSG) for handling non-uniform sampling density; hence, we have three baselines in total, and we train them only with the reconstruction loss.
Please refer to the supplemental material for details of the baseline network architectures.
Although we modify the PointNet, PointNet++, and PointNet++(MSG) architectures for the upsampling problem, for convenience, we still call the baselines by their original names.

Tables~\ref{tab:comparision1} and~\ref{tab:comparision2} list the quantitative comparison results on our collected dataset and the SHREC15 dataset, respectively.
%The unit of deviation mean and std is $10^{-2}$.
Note that measuring NUC with small $p$ shows local distribution uniformity in small regions, while measuring NUC with large $p$ shows more global uniformity.
Among the baselines, PointNet performs the worst, since it cannot capture local structure information. 
Compared with PointNet++, PointNet++(MSG) can slightly improve the uniformity due to the explicit multi-scale information grouping.  
However, it involves more parameters, thus significantly prolonging the training and testing time.
Overall, our PU-Net achieves the best performance with the lowest deviation from surface and the best distribution uniformity compared to the baselines, especially on the local uniformity.
%, which demonstrates the effectiveness of our network.

\begin{figure*}[t]
	\centering
	\includegraphics[width=0.98\linewidth]{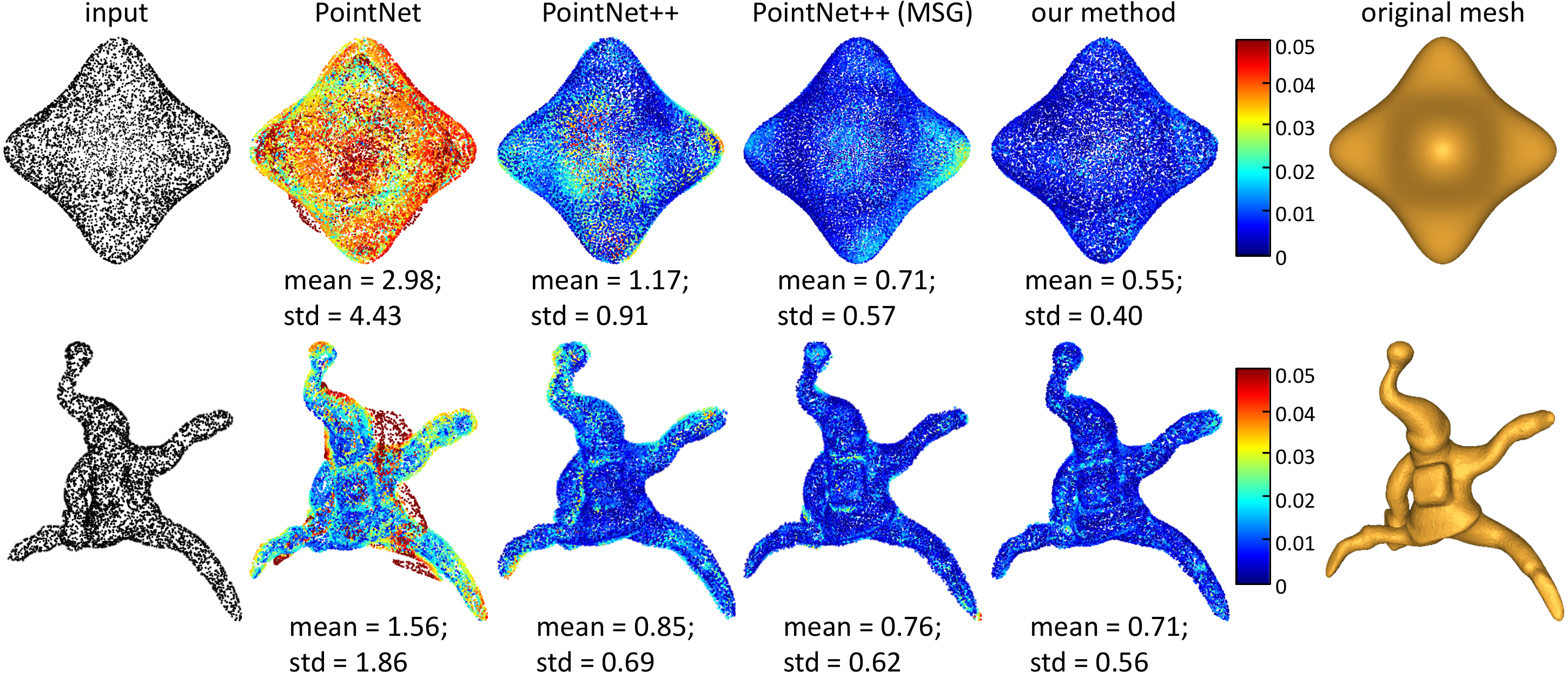}\vspace{1mm}
	\caption{Visual comparison on samples from our collected dataset (top row) and SHREC (bottom row).
The colors on points (see color map) reveal the surface distance errors, where blue indicates low error and red indicates high error.}
	\label{fig:visualcomp}\vspace{-2mm}
\end{figure*}

Fig.~\ref{fig:visualcomp} shows results for visual comparison, where points are colored by their distance deviations from surface.
As we can see, the point clouds predicted by our method better match the underlying surface with lower deviations.  

%%%%%%%%%%%%%%%%%%%%%%%%%%%%%%%%%%%%%%%%%%%%%%%%%%%%%%%%%%%%
\subsection{Architecture Design Analysis}
\label{sec:archtanalysis}

\para{Analyzing the feature expansion.} 
We compare our proposed feature expansion scheme with two interpolation-like schemes.  
The first one is similar to a naive point interpolation (denoted as \emph{interp1}). 
After extracting the point feature of each point, we combine its own features and the features from the nearest neighboring points to generate the upsampled features. 
The second one introduces more randomness (denoted as \emph{interp2}). 
Instead of using the features from the nearest neighbors, we use a radius based ball query to find the neighborhood and combine the features from these points to generate the upsampled features. 
We train these two networks with the reconstruction loss (also named as the EMD loss) and the results are listed in the top two rows of Table~\ref{tab:ablation}.
For fair comparison, we also train our network only with the EMD loss.
The results are shown in the fourth row.  
Comparing with these two interpolation-like schemes, our proposed scheme can generate more uniform outputs with comparable surface distance deviation.

\para{Comparing different loss functions.} 
As mentioned above, the EMD can better capture the object shape than CD.
Comparing their performance in Table~\ref{tab:ablation}, we can see that the EMD loss improves the output uniformity with low surface distance deviation when comparing with the CD loss, meaning that the EMD loss can better encourage the output points to lie on the underlying surface. 
Furthermore, by comparing the last two rows in Table~\ref{tab:ablation}, we can see that the repulsion loss can further improve the uniformity of the output. 

\begin{table}
	\centering
	\caption{Architecture design analysis on our collect dataset.}
	\label{tab:ablation}
	\resizebox{0.5\textwidth}{!}{
		\begin{tabular}{l|c|c|c|c||c|c} \toprule[1pt]
			\multirow{2}*{Methods}	& \multicolumn{4}{c||}{NUC with different $p$} & \multicolumn{2}{c}{Deviation ($10^{-2}$)} \\
			\cline{2-7} 
			&0.4\%	&0.6\%	&0.8\% 	&1.0\%	 & mean	& std	\\ \hline \hline
			\emph{interp1}+EMD  	&0.153	&0.136	&0.126	&0.121		&0.67	&0.54	\\ \hline
			\emph{interp2}+EMD  	&0.144	&0.129	&0.122	&0.118	&0.71	&0.57	\\ \hline
			CD  				  	&0.185	&0.167	&0.154	&0.147	&0.87	&0.69	 \\ \hline
			EMD					  	&0.140	&0.126	&0.119	&0.116	&0.68	&0.58\\ \hline
			Ours			&\BE{0.138}&\BE{0.122}&\BE{0.115}&\BE{0.112}&\BE{0.63}	&\BE{0.53}\\ \bottomrule[1pt]	
		\end{tabular}
	}
	\vspace{-5.5mm}
\end{table}

%%%%%%%%%%%%%%%%%%%%%%%%%%%%%%%%%%%%%%%%%%%%%%%%%%%%%%%%%%%%%%%%%%%%%%%%%%%%%%%%%%%%%
\begin{figure*}[t]
	\centering
	\includegraphics[width=0.96\linewidth]{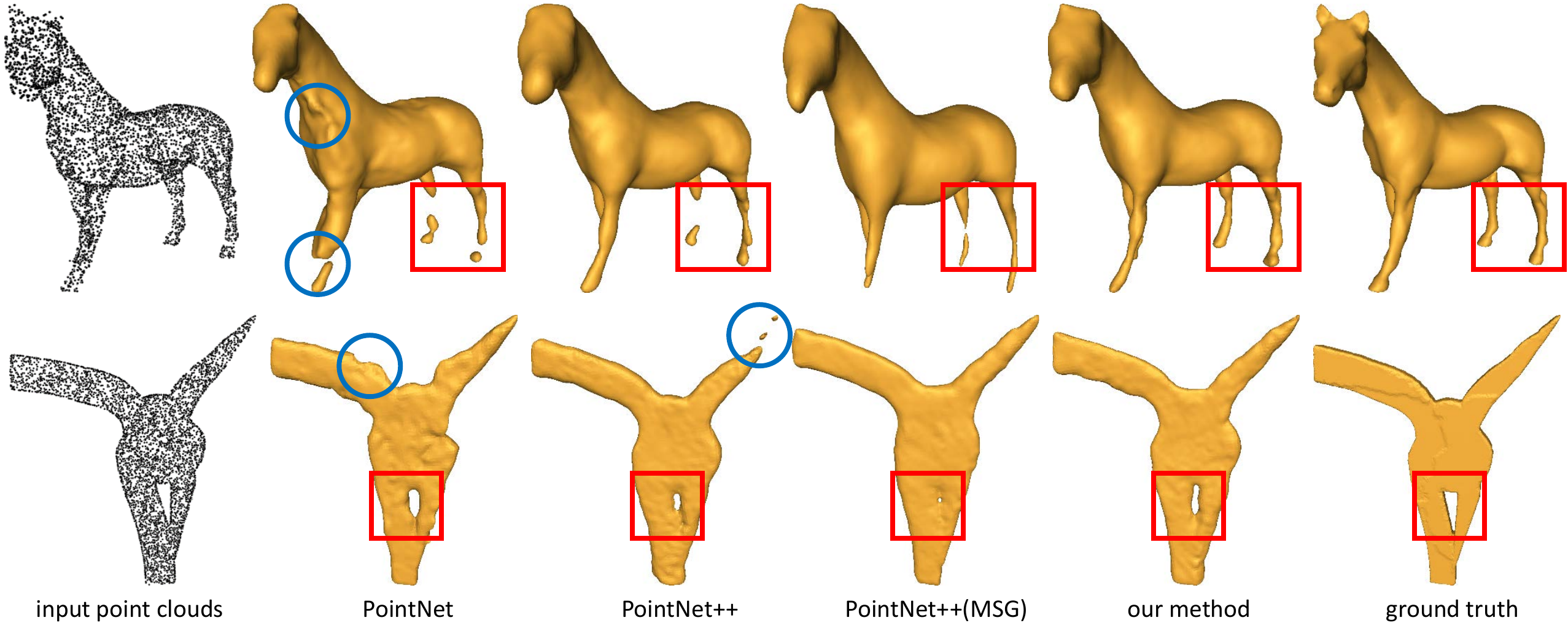}\vspace{1mm}
	\vspace{-2mm}
	\caption{Surface reconstruction results from the upsampled point clouds.}
	\label{fig:surfacereconstruction}
	\vspace{-3.5mm}
\end{figure*} 

\begin{figure}[htbp]
	\centering
	\includegraphics[width=0.99\linewidth]{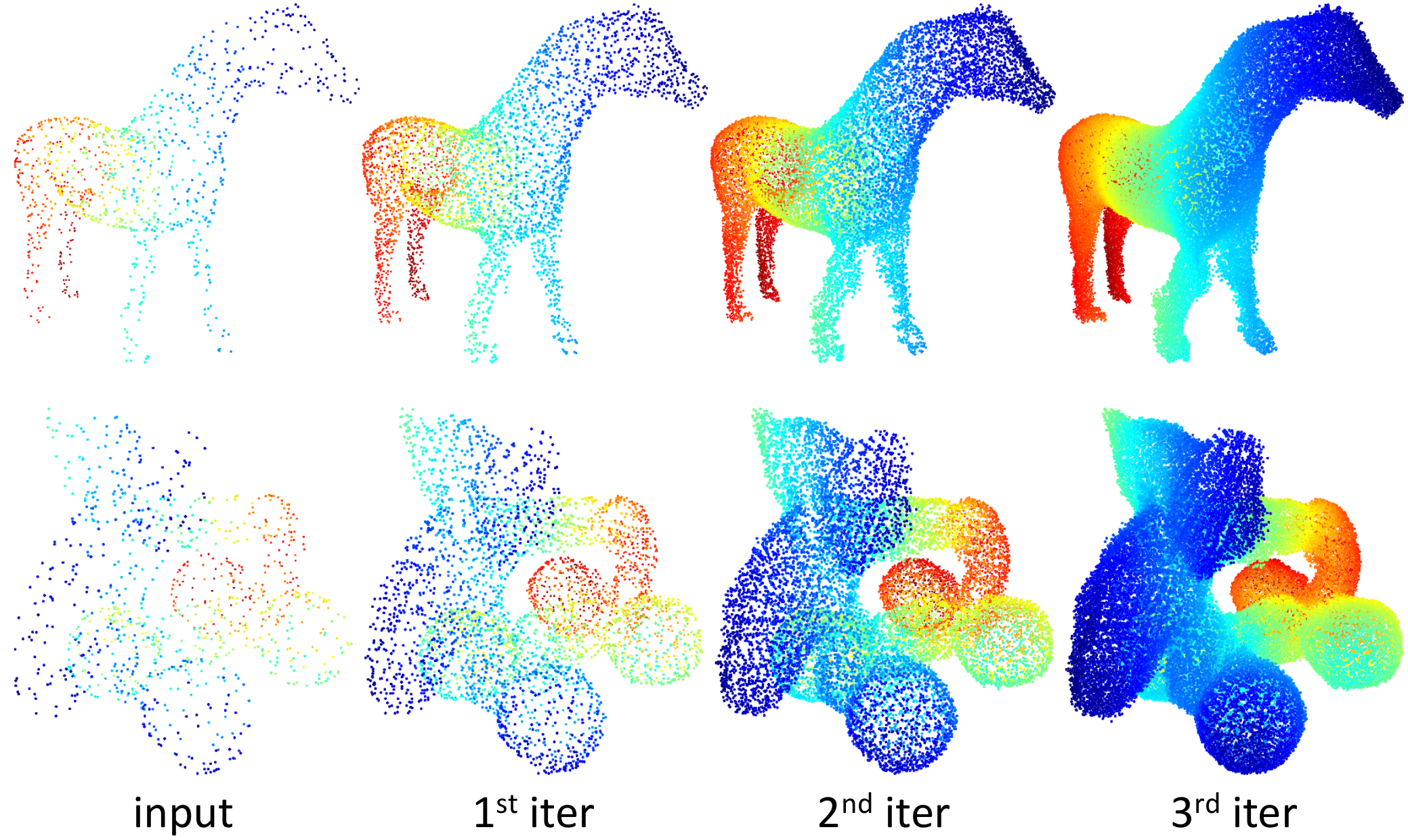}\vspace{1mm}
	\vspace{-1mm}
	\caption{Results of iterative upsampling. We color-code points by the depth information. Blue points are closer to us.}
	\label{fig:iterative}
%\vspace{-2mm}
\end{figure}

\begin{figure*}[!t]
	\centering
	\includegraphics[width=13.5cm]{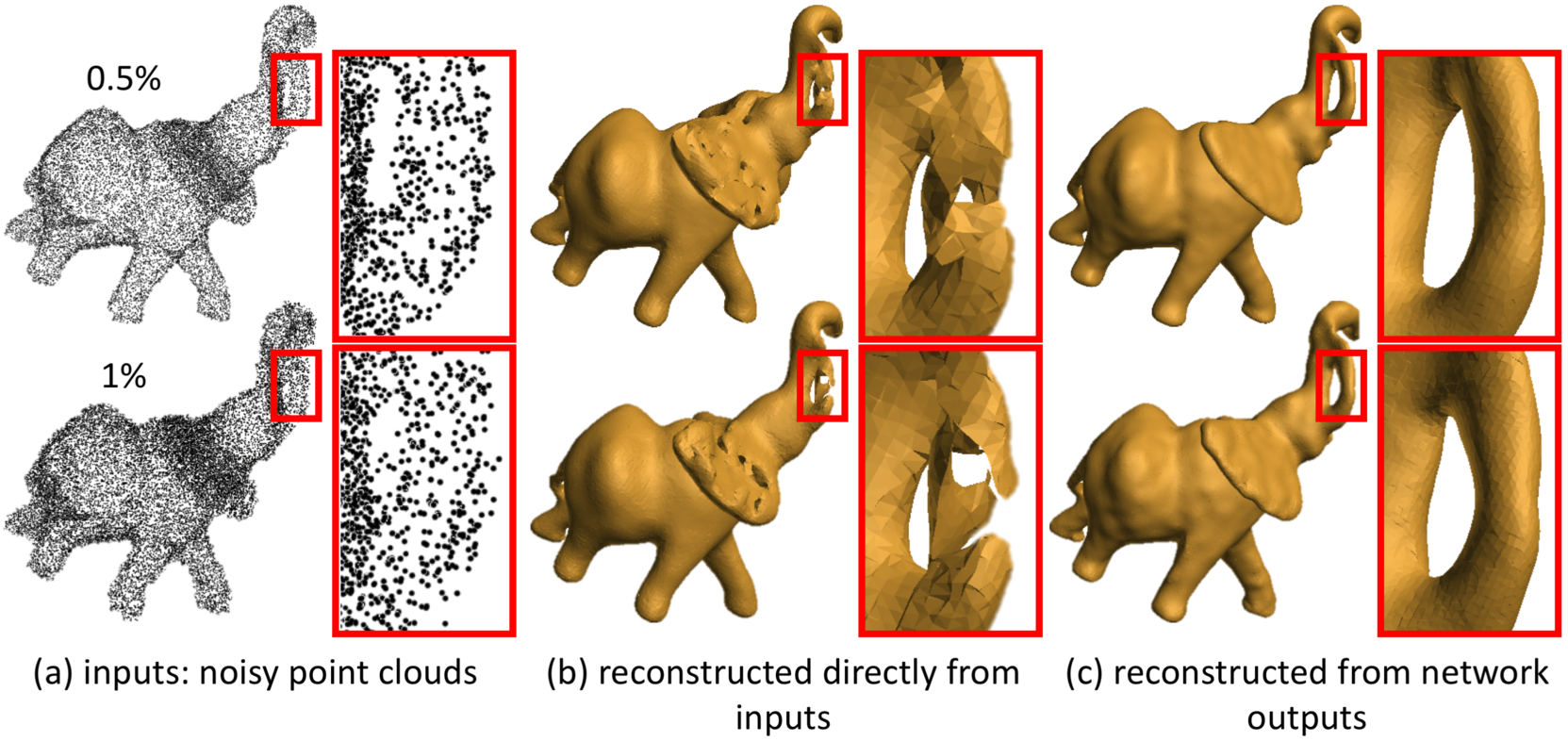}
	\caption{Surface reconstruction results from noisy input points.}
	\label{fig:noise}\vspace{-2mm}
\end{figure*}

\subsection{More Experiments}
\label{sec:moreresults}

\para{Surface reconstruction from upsampled point sets.} 
An important application of point cloud upsampling is to improve the surface reconstruction quality.
Hence, we compare the reconstruction results of different methods with the direct Poisson surface reconstruction method~\cite{poisson2006} provided in MeshLab~\cite{LocalChapterEvents:ItalChap:ItalianChapConf2008:129-136}; see Fig.~\ref{fig:surfacereconstruction}. 
We can observe that the reconstruction result from our method is the closest to the ground truth, while other methods either miss certain structures (\eg, the leg of the Horse) or overfill the hole.

\para{Results of iterative upsampling.}   
To study the ability of our network to handle varying number of input points, we design an iterative upsampling experiment, which takes the output of the previous iteration as the input of the next iteration.
Fig.~\ref{fig:iterative} shows the results.
The initial input point cloud has 1024 points and we increase fourfold in each iteration.
From the results, we can see that our network can produce reasonable results for different number of input points.
Furthermore, this iterative upsampling experiment also shows the anti-noise ability of our network to resist the accumulated errors introduced in the iterative upsampling.

\para{Results from noisy input point sets.}
Fig.~\ref{fig:noise} shows the surface reconstruction results from noisy point clouds (Gaussian noise of 0.5\% and 1\% of object bounding box diagonal), which demonstrate that our network facilitates the production of better surfaces even with noisy inputs.

\begin{figure}[h]
	\centering
	\includegraphics[width=0.9\linewidth]{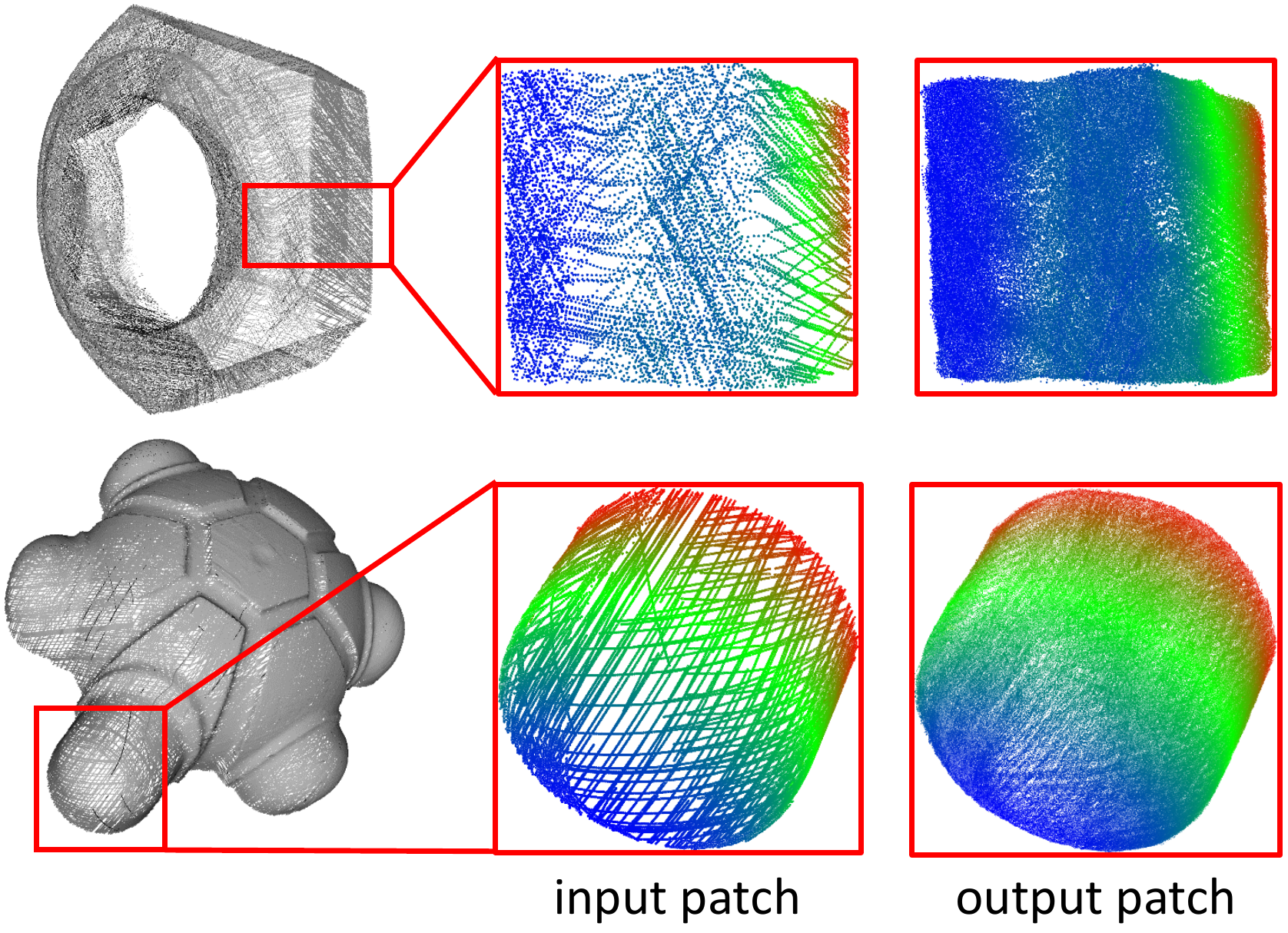}\vspace{1mm}
	\caption{Results on real-scanned point clouds (Screw nut \& Turtle). We color-code input patches and upsampling results to show the depth information. Blue points are closer to viewpoint.}
	\label{fig:realscan}\vspace{-2mm}
\end{figure}

\para{Results on real-scanned point clouds.}
Lastly, we evaluated the ability of our network to upsample real-scanned point clouds, which were downloaded from Visionair~\cite{timmurphy.org}.
In Fig.~\ref{fig:realscan}, the left-most column presents the real-scanned point clouds. 
Even though each real-scanned point cloud contains millions of points, the phenomenon of inhomogeneity still exists.
For better visualization, we cut small patches from the original point clouds and show the patches in the middle column.
We can observe that the real-scanned points tend to have line structures, while our network still has the ability to uniformly add points in the sparse regions.

%% file: conclusion.tex
\section{Conclusion}
\label{sec:conclusion}
In this paper, we present a deep network for point cloud upsampling, with the goal of generating a denser and uniform set of points from a sparser set of points.
Our network is trained at a patch-level using a multi-level feature aggregation manner, thus capturing both local and global information.
The design of our network bypasses the need for a prescribed order among the points, by operating on individual features that contain non-local geometry to allow a context-aware upsampling. Our experiments demonstrate the effectiveness of our method.
As the first attempt using deep networks, our method still has a number of limitations.
Firstly, it is not designed for completion, so our network can not fill large holes and missing parts. 
Besides, our network may not be able to add meaningful points for tiny structures that are severely undersampled.

In the future, we would like to investigate and develop more means to handle irregular and sparse data, both for regression purposes and for synthesis. One immediate step is to develop a downsampling method. Although, downsampling seems like a simpler problem, there is room to devise proper losses and architecture that maximize the preservation of information in the decimated point set. We believe that in general, the development of deep learning methods for irregular structures is a viable research direction.

%% file: supp.tex
\appendix
\onecolumn
%\section*{Supplementary}

%%%%%%%%%%%%%%%%%%%%%%%%%%%%%%%%%%%%%%%%%%%%%%%%%%%%%%%%%%%%%%%%%%%%%%%%%%%%%%
\section{Overview}

In this supplementary material, we first provide more details about our collected dataset in Section~\ref{supp_dataset}.
Then, we show the details of our network architecture as well as the baseline networks employed in the experiments in Section~\ref{supp_network}.
%Finally, we provide visualization results on ModelNet40 and ShapeNet tested by our network in Section~\ref{supp_result}.

%%%%%%%%%%%%%%%%%%%%%%%%%%%%%%%%%%%%%%%%%%%%%%%%%%%%%%%%%%%%%%%%%%%%%%%%%%%%%%

\section{Details of our Collected Dataset}
\label{supp_dataset}

\begin{figure*}[h]
	\begin{adjustwidth}{0.5cm}{0.5cm}
		\begin{center}
			\small
			\setlength{\tabcolsep}{3pt}
			\begin{tabular}{  c  c  c  c  c  c  }
				{\graphicspath{{figs/figDRCN/}}\includegraphics[height=0.2\textwidth]{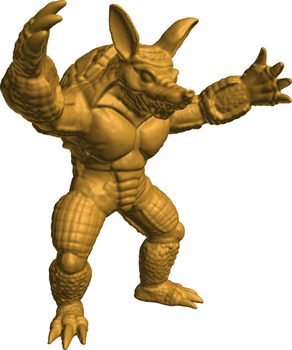}}
				& {\graphicspath{{figs/figDRCN/}}\includegraphics[height=0.18\textwidth]{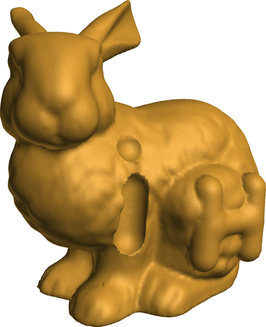}}
				& {\graphicspath{{figs/figDRCN/}}\includegraphics[height=0.2\textwidth]{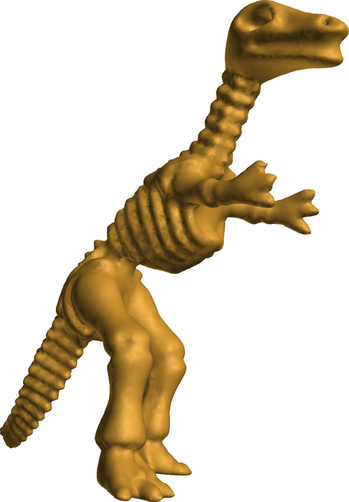}}
				& {\graphicspath{{figs/figDRCN/}}\includegraphics[height=0.2\textwidth]{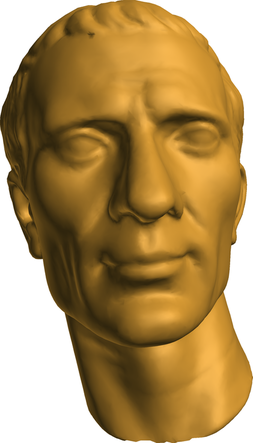}}
				& {\graphicspath{{figs/figDRCN/}}\includegraphics[height=0.2\textwidth]{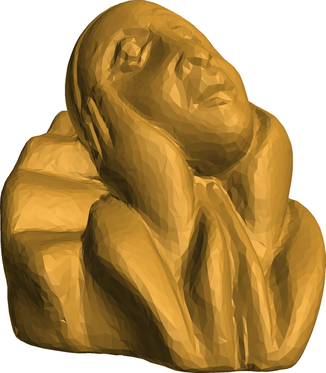}}
				& {\graphicspath{{figs/figDRCN/}}\includegraphics[height=0.2\textwidth]{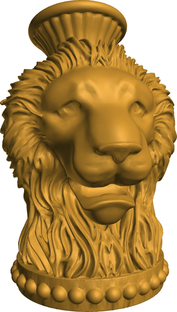}}
				\\
				Armadillo	&Bunny		&Dino		&Julius		&Pierrot		&Vaselion\\
				
				{\graphicspath{{figs/figDRCN/}}\includegraphics[width=0.15\textwidth]{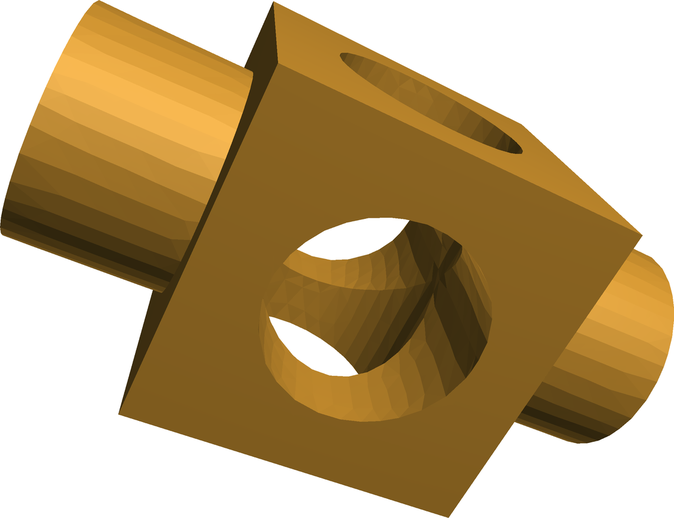}}
				& {\graphicspath{{figs/figDRCN/}}\includegraphics[width=0.15\textwidth]{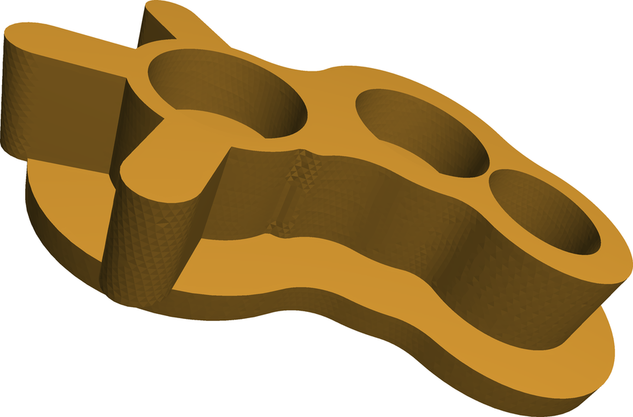}}
				& {\graphicspath{{figs/figDRCN/}}\includegraphics[width=0.15\textwidth]{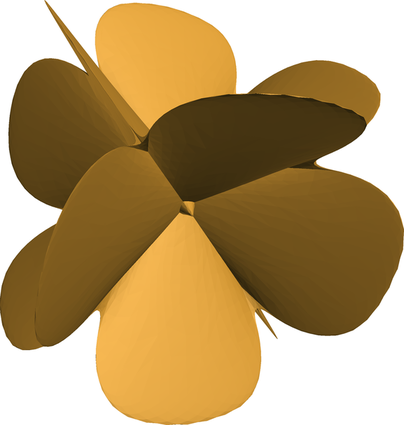}}
				& {\graphicspath{{figs/figDRCN/}}\includegraphics[width=0.15\textwidth]{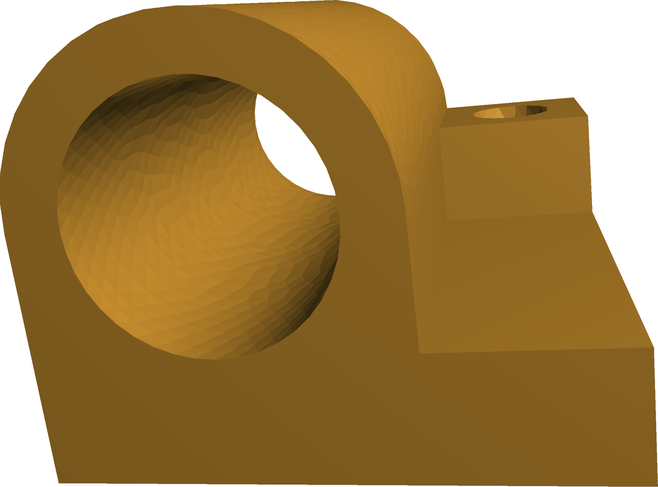}}
				& {\graphicspath{{figs/figDRCN/}}\includegraphics[width=0.12\textwidth]{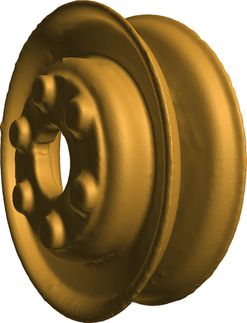}}
				& {\graphicspath{{figs/figDRCN/}}\includegraphics[width=0.15\textwidth]{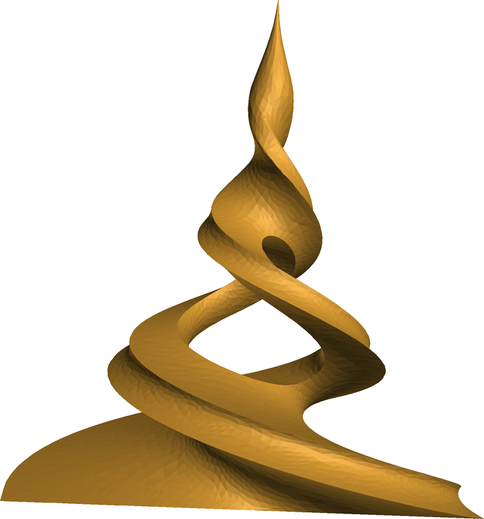}}
				\\
				Block	&Cad		&Focal\_octa		&Joint		&Pulley		&Twirl\\	
			\end{tabular}
			\caption{Examples 3D models in our training dataset.
			The first row shows 3D models with smooth surface regions, while the second row shows 3D models with sharp corners and edges.}
			\label{fig:dataset_train}
		\end{center}
	\end{adjustwidth}
\end{figure*}

\begin{figure*}[h]
	\begin{adjustwidth}{0.5cm}{0.5cm}
		\begin{center}
			\small
			\setlength{\tabcolsep}{3pt}
			\begin{tabular}{  c  c  c  c  c  c  }
				{\graphicspath{{figs/figDRCN/}}\includegraphics[width=0.12\textwidth]{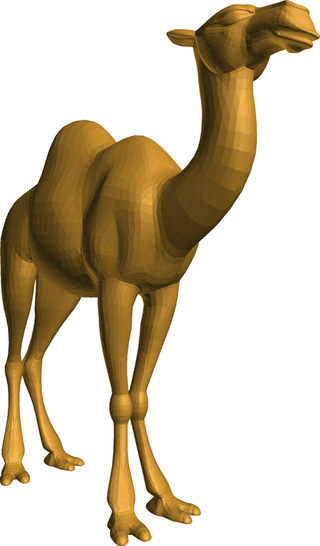}}
				& {\graphicspath{{figs/figDRCN/}}\includegraphics[width=0.15\textwidth]{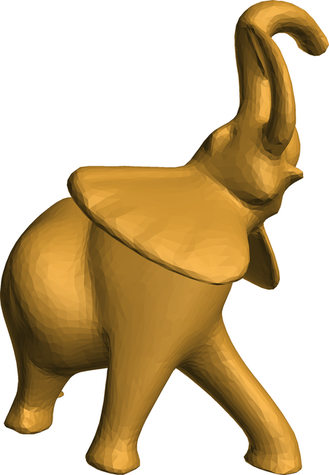}}
				& {\graphicspath{{figs/figDRCN/}}\includegraphics[width=0.15\textwidth]{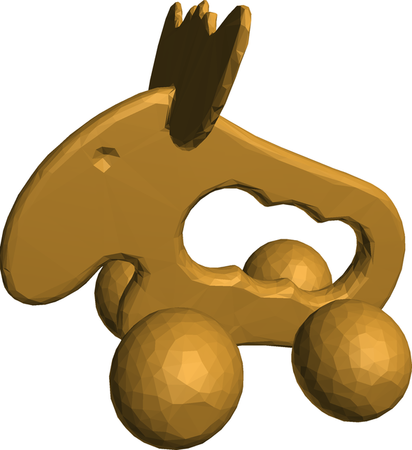}}
				& {\graphicspath{{figs/figDRCN/}}\includegraphics[width=0.15\textwidth]{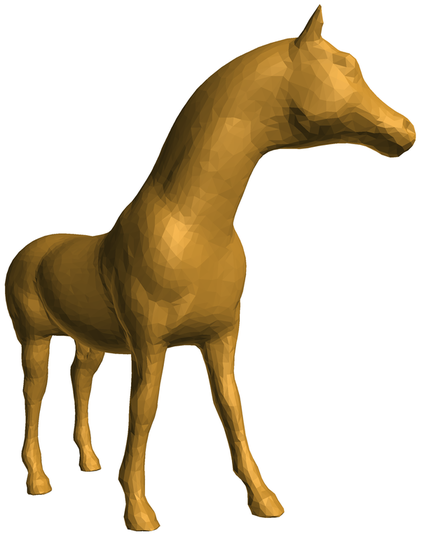}}
				& {\graphicspath{{figs/figDRCN/}}\includegraphics[width=0.12\textwidth]{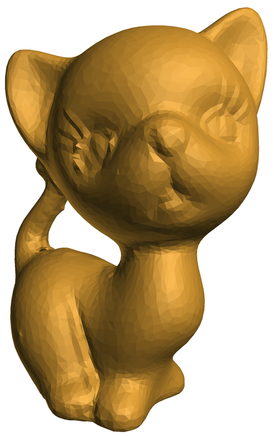}}
				& {\graphicspath{{figs/figDRCN/}}\includegraphics[width=0.10\textwidth]{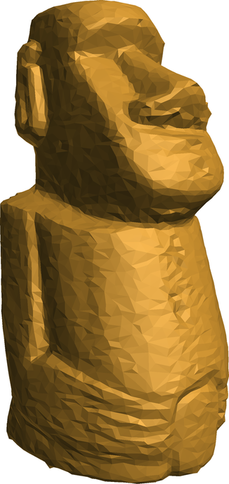}}
				\\
				Camel	&Elephant		&Elk		&Horse		&Kitten		&Moai\\
				
				{\graphicspath{{figs/figDRCN/}}\includegraphics[width=0.15\textwidth]{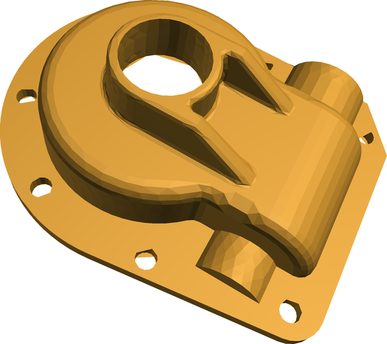}}
				& {\graphicspath{{figs/figDRCN/}}\includegraphics[width=0.10\textwidth]{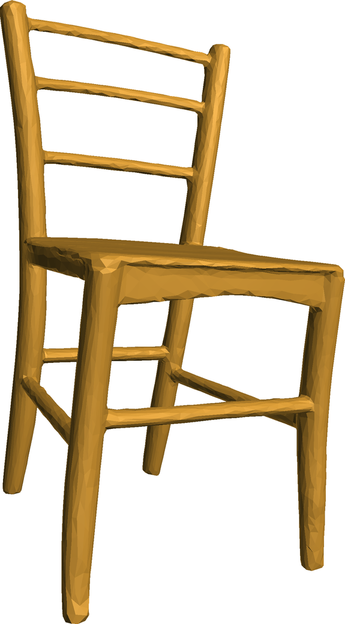}}
				& {\graphicspath{{figs/figDRCN/}}\includegraphics[width=0.15\textwidth]{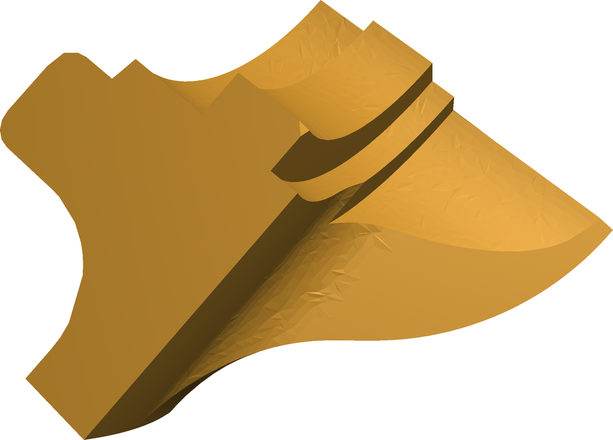}}
				& {\graphicspath{{figs/figDRCN/}}\includegraphics[width=0.15\textwidth]{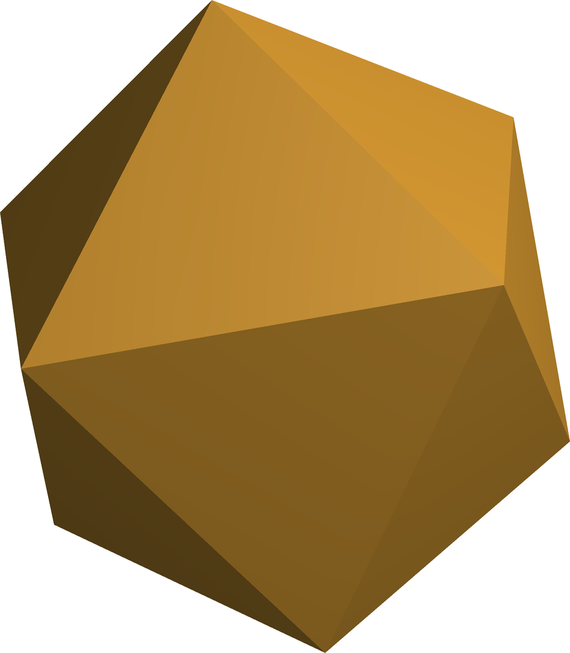}}
				& {\graphicspath{{figs/figDRCN/}}\includegraphics[width=0.15\textwidth]{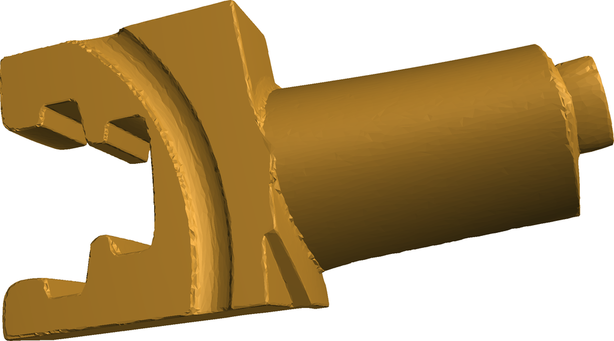}}
				& {\graphicspath{{figs/figDRCN/}}\includegraphics[width=0.15\textwidth]{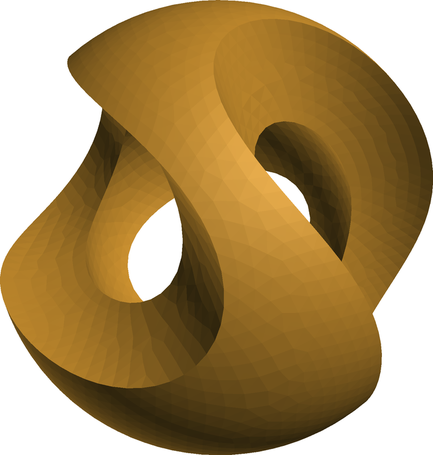}}
				\\
				Casting	&Chair		&Fandisk		&Icosahedron		&Quadric		&Sculpt\\

			\end{tabular}
			\caption{Examples 3D models in our testing dataset.
			The first row shows 3D models with smooth surface regions, while the second row shows 3D models with sharp corners and edges.}
			\label{fig:dataset_test}
		\end{center}
	\end{adjustwidth}
\end{figure*}

We collect 60 different 3D models to form our training and testing datasets.
The specific name of each model is shown in Table~\ref{tab:dataset}.
We also present the shapes of some training and testing 3D models in our dataset in Fig.~\ref{fig:dataset_train} and Fig.~\ref{fig:dataset_test}, respectively.  
As we can see, our collected datasets have a large variation in geometry shapes, containing 3D models with smooth surface regions (first row) and 3D models with sharp corners and edges (second row). 
There is also a large variation between training and testing 3D models, indicating a good generalization ability of our proposed method. 

\begin{table*}[h]
	\caption{The complete name list of the 3D models in our training and testing datasets.}
	\label{tab:dataset}
	\centering
	\begin{center}
		\begin{tabular}{|c|m{30em}|} \toprule[1pt]
			&{\qquad \qquad \qquad \qquad \qquad Model Names}\\ \hline 
			Training & 	Armadillo, Boy1, Boy2, Bumpy\_torus, Bunny, Cad, Cylinder, Child1, Child2, Chinese\_lion, 
						Cone, Cup, Dino, Egea, Ellipsoid, Eros, Fish, Focal\_octa, Gargoyle, Girl1, Girl2, Hand, Joint, 
						Julius, Nicolo, Octa\_flower, Pierrot, Pulley, Pyramid1, Pyramid2, Retinal, Rolling\_stage, 
						Screwdriver, Sharp\_sphere, Special\_cube, Star1, Turbine, Twirl, Vaselion  \\ \hline

			Testing  &	Camel, Casting, Chair, Cover\_rear, Cow, Duck, Eight, Elephant, Elk, Fandisk, Genus, Horse, 
						Icosahedron, Kitten, Moai, Octahedron, Pig, Quadric, Sculpt, Star2\\ \bottomrule[1pt]
		\end{tabular}
	\end{center}
\end{table*}

\clearpage

\clearpage

%%%%%%%%%%%%%%%%%%%%%%%%%%%%%%%%%%%%%%%%%%%%%%%%%%%%%%%%%%%%%%%%%%%%%%%%%%%%%%

\section{Details of Network Architectures}
\label{supp_network}

The details of our network architecture are listed as follows.
\begin{itemize}
	\item
	In the hierarchical feature learning component, we use four levels to extract local features. Following the notations in PointNet++, we use ($K$, $r$, [$l_1,...,l_d$]) to represent a level with $K$ local regions of ball radius $r$, and [$l_1,...,l_d$] the $d$ fully connected layers with width $l_i$ ($i=1,...,d$). Therefore, the parameters we use are $(N,0.05,[32,32,64])$, $(N/2,0.1,[64,64,128])$, $(N/4,0.2,[128,128,256])$ and $(N/8, 0.3,[256,256, 512])$.
	\item
	In the multi-level feature aggregation component, we use interpolation to restore the feature of each level and use a convolution to reduce the restored feature to 64 dimensions. Therefore, $\tilde{C}$ = 259 in our experiments.
	\item 
	In the feature expansion component, the output feature channel numbers $\tilde{C}_1$ and $\tilde{C}_2$ are 256 and 128, respectively.
	\item 
	In the coordinate reconstruction component, we use two fully connected layers with 64 and 3 output channels, respectively.
\end{itemize}

\begin{figure*}
	\begin{center}
		\includegraphics[width=0.6\linewidth]{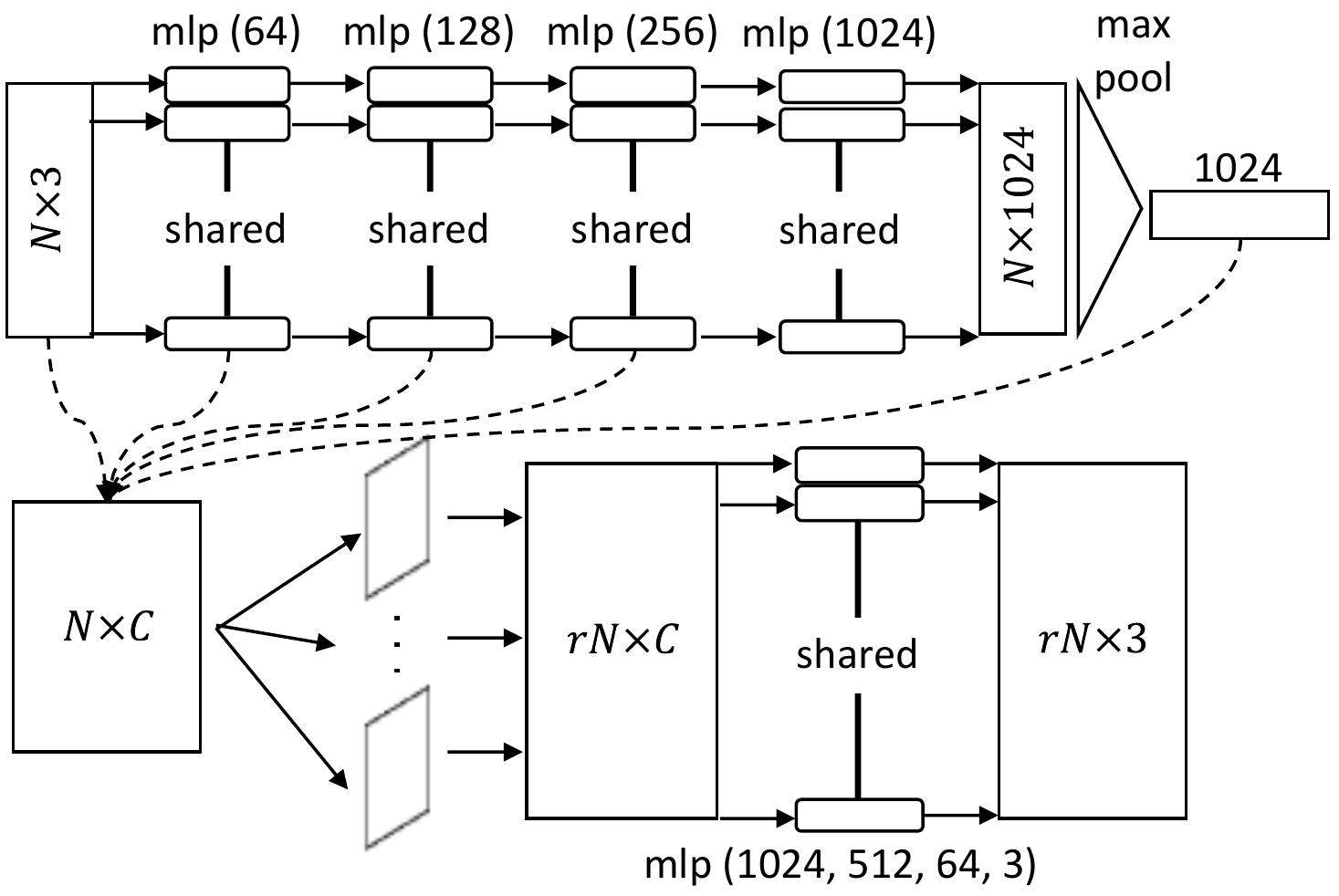}
	\end{center}  
	\caption{The network architecture of PointNet for point cloud upsampling.}
	\label{fig:pointnet}
\end{figure*}

\begin{figure*}
	\begin{center}
		\includegraphics[width=0.8\linewidth]{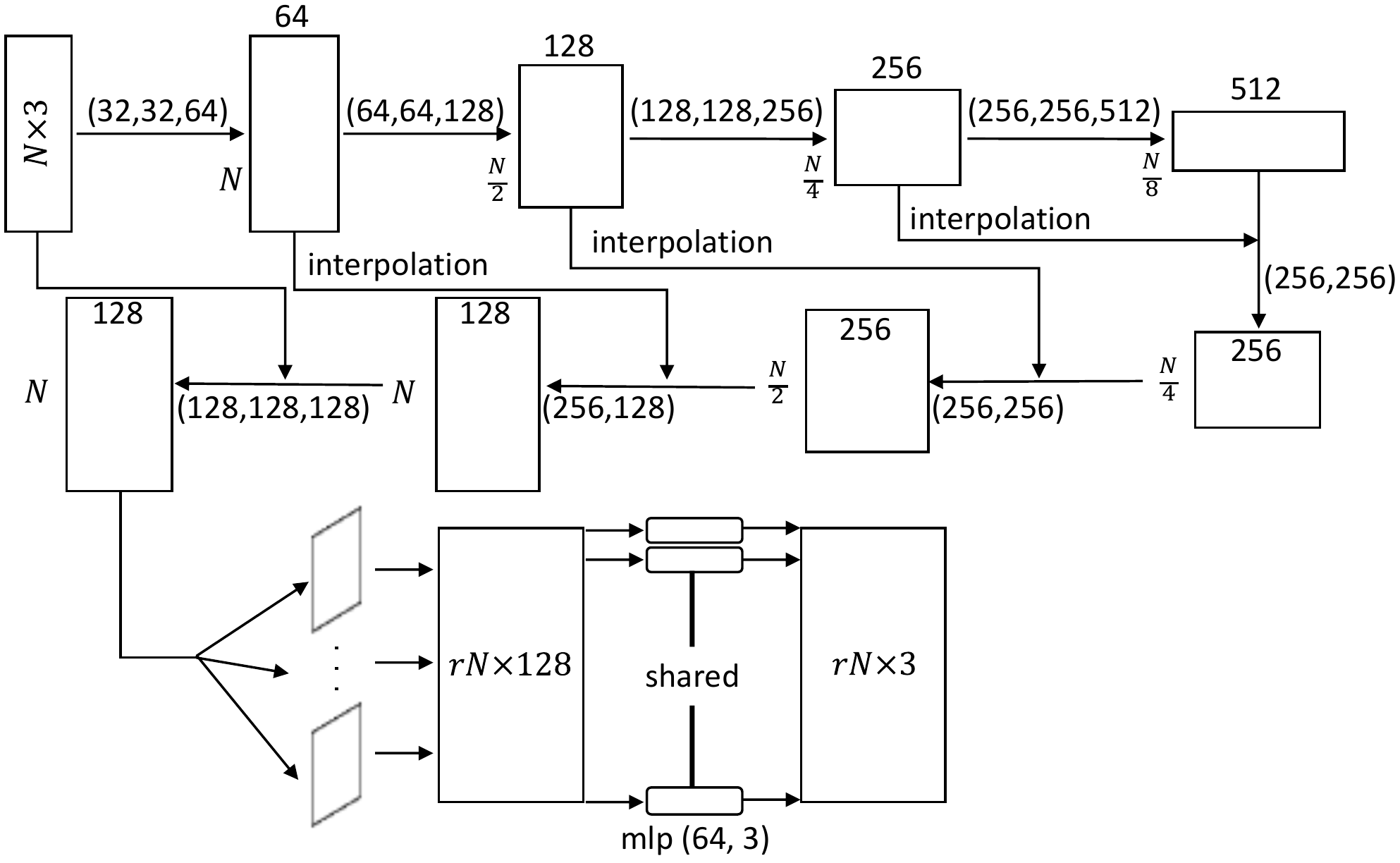}
	\end{center} 
	\caption{The network architecture of PointNet++ for point cloud upsampling.}
	\label{fig:pointnet2}
\end{figure*}

\begin{figure*}[p]
	\begin{center}
		\includegraphics[width=0.8\linewidth]{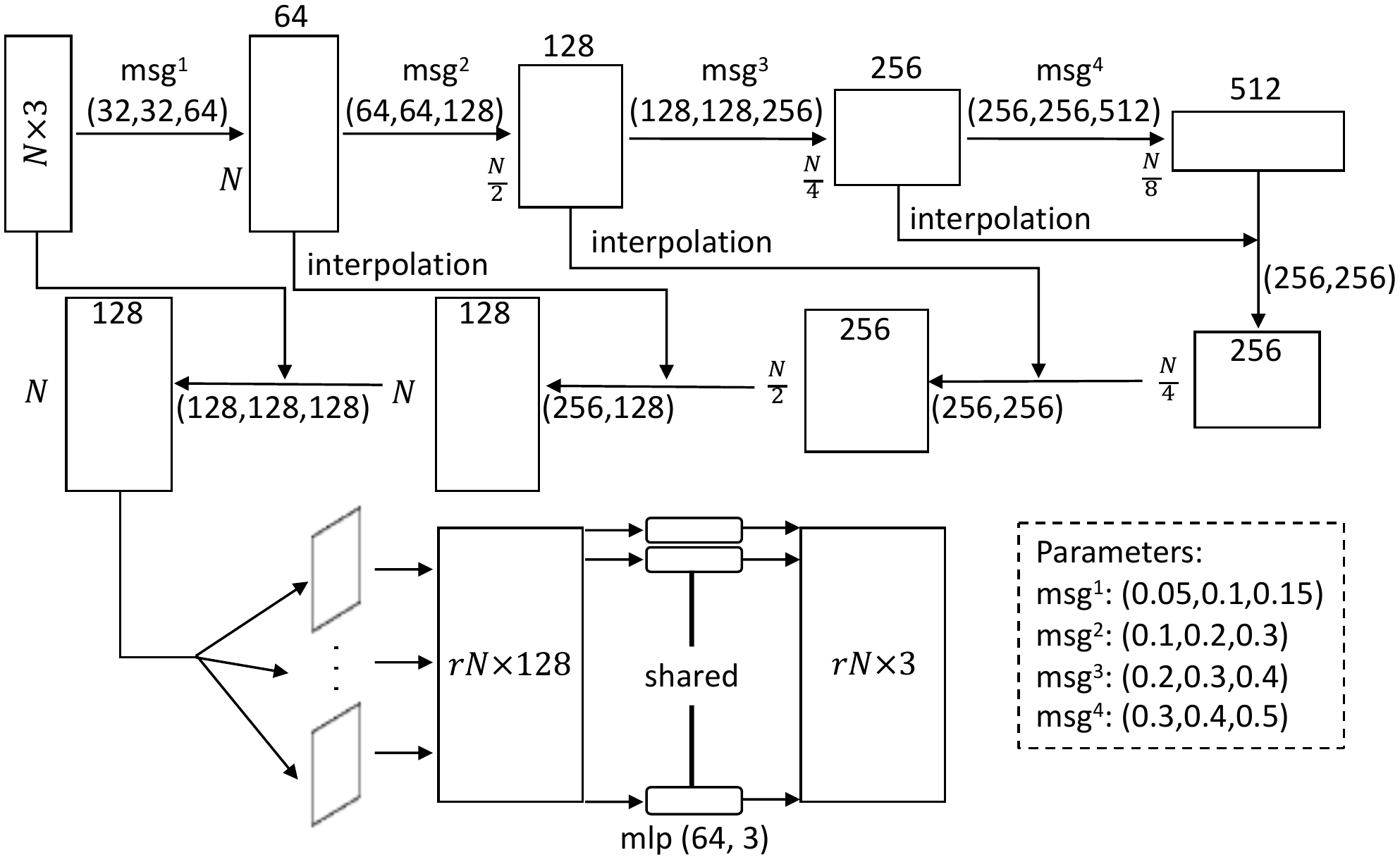}
	\end{center} 
	\caption{The network architecture of PointNet++(MSG) for point cloud upsampling.}
	\label{fig:pointnet2_msg}
\end{figure*}

The details of the baseline architectures are illustrated in Fig.~\ref{fig:pointnet}, Fig.~\ref{fig:pointnet2} and Fig.~\ref{fig:pointnet2_msg}.

All the convolution layers and fully connected layers in the above networks are followed by the ReLU operator, except for the last coordinate regression layer.

\clearpage

\clearpage